\documentclass{article} 
\usepackage{iclr2023_conference_tinypaper,times}

\usepackage{amsmath,amsfonts,bm}

\def\eqref#1{equation~\ref{#1}}

\def\1{\bm{1}}

\DeclareMathAlphabet{\mathsfit}{\encodingdefault}{\sfdefault}{m}{sl}
\SetMathAlphabet{\mathsfit}{bold}{\encodingdefault}{\sfdefault}{bx}{n}

\usepackage{hyperref}
\usepackage{url}

\usepackage{amsfonts} 
\usepackage{booktabs} 
\usepackage{multirow} 
\usepackage{subfigure} 
\usepackage[utf8]{inputenc}
\usepackage{mathtools}
\usepackage{amssymb}
\usepackage{amsthm}
\usepackage{amsmath}
\usepackage{enumitem}
\usepackage{graphicx}
\usepackage{cancel}
\usepackage{color,xcolor}
\usepackage{colortbl}
\usepackage{hyperref}
\usepackage{tabu}
\usepackage{array}
\usepackage{siunitx}
\usepackage{booktabs}
\usepackage{makecell}
\definecolor{Gray}{gray}{0.9}
\usepackage{inconsolata}

\title{Can Perplexity Reflect Large Language \\Model's Ability in Long Text Understanding?}

\author{Yutong Hu, Quzhe Huang, Mingxu Tao, Chen Zhang, Yansong Feng \thanks{Corresponding author} \\
Wangxuan Institute of Computer Technology, Peking University \\
\texttt{\{huyutong,huangquzhe,thomastao,zhangch,fengyansong\}@pku.edu.cn} \\
}

\iclrfinalcopy 
\begin{document}

\maketitle

\begin{abstract}
Recent studies have shown that Large Language Models (LLMs) have the potential to process extremely long text. 
Many works only evaluate LLMs' long-text processing ability on the language modeling task, with perplexity (PPL) as the evaluation metric. However, in our study, we find that there is no correlation between PPL and LLMs' long-text understanding ability. Besides, PPL may only reflect the model's ability to model local information instead of catching long-range dependency. Therefore, only using PPL to prove the model could process long text is inappropriate. The local focus feature of PPL could also explain some existing phenomena, such as the great extrapolation ability of the position method ALiBi. When evaluating a model's ability in long text, we might pay more attention to PPL's limitation and avoid overly relying on it.





\end{abstract}

\section{INTRODUCTION}

Recently, many researchers~\citep{chen2023extending,chen2023clex,xiong2023effective, ding2023longnet, chen2023longlora} have proposed various approaches to scale up the context window of LLMs to more than 100k. 
Since there is not a comprehensive benchmark tailored for the evaluation of such extremely long text understanding, such as question answering (QA) over 100K tokens, researchers use perplexity (PPL), 
an evaluation metric for language modeling~\footnote{The definition and calculation method of PPL is shown in Appendix~\ref{app:ppl}. A lower PPL shows a higher accuracy of a model in long-text language modeling}, to demonstrate the model's ability to process long text~\citep{chen2023longlora, ding2023longnet,liu2023scaling,peng2023yarn}. 

However, only given LLMs are excellent in language modeling, can it indicate LLMs' ability to understand long text? We conduct experiments on three long context window LLM variants to figure out this. We use several available benchmarks of downstream tasks, such as QA and summerization, to evaluate their long-text understanding ability. Surprisingly, the models' performance on language modeling is inconsistent with their performance on most downstream tasks, implying the PPL can not be a good indicator of the model's long-text understanding ability.

We speculate that the phenomenon above may be because PPL is a reflection of the model's ability to model \textbf{local} information. 
We use LLaMA2, which only has a short context window of 4,096 and cannot handle long context, to prove our speculation. The experiment results show that, LLaMA2 delivers comparable PPL with the long context window LLMs. The feature of PPL in reflecting local information modeling ability can also explain why methods such as ALiBi \citep{press2022train}, which makes the model mainly focus on local information, could enable models to extrapolate to longer inference sequences while keeping the PPL at a low level.
\section{Lower PPL $\neq$ Understanding Long Text Better}
\label{sec:sec2}

In this section, we compare the models' performance in language modeling and several downstream tasks, and figure out whether a lower PPL in language modeling indicates a better long-text understanding ability. We choose three model variants whose context windows are longer than 100K tokens for experiments: 1) YARN-7B-128K \citep{peng2023yarn}, 2) Yi-6B-200K \footnote{https://github.com/01-ai/Yi}, 3) LongLoRA-7B-100K \citep{chen2023longlora}. 

We first calculate the PPL of each model with a 76k input length\footnote{76K is the longest input length our machine can process. We get Out Of Memory Error with longer inputs} on 50 books randomly sampled from the test set of PG-19~\citep{rae2019compressive}. While for downstream tasks, we use two public benchmarks, QMSUM~\citep{zhong2021qmsum} and NarrativeQA~\citep{kočiský2017narrativeqa}, to evaluate the models' performance in long question answering and long document summarization. In addition, following \cite{longchat2023}, we use a finer-grained line retrieval test to evaluate models' retrieval ability. The experiment details are shown in Appendix~\ref{app:ppl} and Appendix~\ref{app:tasks}.

The experiment results are shown in Table~\ref{tab:downstream_tasks}. Column {\bf LM} reports models' PPL on the long text. YARN delivers the lowest PPL of 1.878, compared to 2.069 of Yi and 2.002 of LongLoRA, which indicates YARN's better ability in long text language modeling. However, YARN does not deliver the best performance in downstream tasks. Instead, LongLoRA outperforms other models on all downstream tasks. The inconsistency between the models' performance in language modeling and downstream tasks demonstrates that, although PPL is a fair evaluation for language modeling, it can not be a good indicator of long-text understanding ability.

\begin{table*}[t]
    \centering
    \small
    \setlength\tabcolsep{5.7pt}
    \begin{tabular}{c|c|c|c|c|c|c}
        \toprule[2pt]
        {\bf Task} & {\bf LM} & {\bf QMSUM} & {\bf NaQA} & {\bf NaQA (32K-64K)} & {\bf NaQA (64K-128K)} & {\bf Retrieval} \\
        {\bf Metric} & PPL $\downarrow$ & Rouge-L $\uparrow$ & F1 $\uparrow$ & F1 $\uparrow$ & F1 $\uparrow$ & Acc. $\uparrow$ \\
        \midrule
        YARN & {\bf 1.878} & 19.79 & 16.84 & 26.75 & 17.49 & 0.065 \\
        Yi & 2.069 & 20.48 & 14.49 & 23.00 & 21.26 & 0.38 \\
        LongLoRA & 2.002 & {\bf 21.53} & {\bf 18.38} & {\bf 29.23} & {\bf 23.19} & {\bf 0.485} \\
        LLaMA & 1.935 & - & - & - & - & - \\
        \bottomrule[2pt]
    \end{tabular}
    
    \caption{Performance on downstream tasks. LM is language modeling and NaQA is NarrativeQA. 32K-64K (64K-128K) indicates the input length is between 32K and 64K (64K and 128K).}
    \label{tab:downstream_tasks}
\end{table*}

\section{PPL might reflect local language modeling ability}
\label{sec:sec3}
To explain the phenomenon in Sec~\ref{sec:sec2}, we speculate PPL may be related to how well LLMs handle the local information, rather than understanding the whole long text. Specifically, for long text language modeling, LLMs may only use local tokens to predict the next token, finally achieving a low PPL.

To prove the speculation, we calculate the PPL of LLaMA2-7B on the same test corpus as the experiment in Sec~\ref{sec:sec2}. Each token is predicted based on the previous 4k tokens, ensuring LLaMA2 can only use the local information in language modeling. We show experiment details in Appendix~\ref{app:ppl}. 

The experiment results in Table~\ref{tab:downstream_tasks} show that LLaMA2 delivers a PPL of 1.935, lower than Yi and LongLoRA with an input length of 76K, demonstrating that a model incapable of understanding long text can also deliver a low PPL. In other words, the good performance of a model in long-text language modeling can not indicate the model's ability in long-text understanding. Therefore, PPL may only reflect LLMs' ability to handle the local information, rather than understanding long text. 

As LLMs can only use local information to deliver low PPL on long text, it can explain the great extrapolation ability of position embedding ALiBi~\citep{press2022train}. Since the position representation method of ALiBi penalizes attention scores between distant query-key pairs, LLMs still focus on the local information even with longer input, which is why the PPL can remain at a similar level.

\section{CONCLUSION}

PPL can be an effective evaluation metric for long-text language modeling ability, but not for long-text understanding. A model without the ability to understand long text can also effectively use local information to model a long text. Considering PPL can not be a good indicator for long text understanding ability, except for using PPL to evaluate a model, we call for more diversified evaluation metrics for long text processing ability from multiple aspects.


\subsubsection*{URM Statement}
The authors acknowledge that at least one key author of this work meets the URM criteria of ICLR 2024 Tiny Papers Track.

\bibliography{iclr2023_conference_tinypaper}
\bibliographystyle{iclr2023_conference_tinypaper}

\appendix

\clearpage
\section{Appendix}

\subsection{PPL Calculation Method}
\label{app:ppl}
For the method we use for LongLoRA, Yi and YARN in Sec~\ref{sec:sec2}, we split the given text into $N$ chunks with the size of 76k per chunk. Taking each chunk as input, the LLM makes predictions on every token based on all of its previous tokens in the chunk. Specifically, the probability of token $x_i$ in the chunk is calculated as:
\begin{equation}
    P_1(x_i) = p(x_i|x_0, ..., x_{i-1}), i \in [0, 76000)
\end{equation}
After obtaining the prediction results of all chunks, i.e., obtaining the probability of every token in the given text, we calculate the 
\begin{equation}
    PPL = average(-log(P_1(x_i))), i \in [0, L)
\end{equation}
where $L$ is the length of the given text. The calculation method is the same as~\citet{chen2023longlora, peng2023yarn}.

In the method we use for LLaMA2 in Sec~\ref{sec:sec3}, every token is predicted based on a fixed number of previous tokens. This is different from the former method, which makes predictions on the $i_{th}$ token based on all previous tokens in the chunk. Specifically, if the index of the token in the given text is equal to or larger than the context size of 4k, the probability of the token is calculated based on its corresponding previous 4k tokens. Otherwise, its probability is calculated based on all of its previous tokens. The equations are: 
\begin{equation}
    P_2(x_i) = p(x_i|x_{i-4000}, ..., x_{i-1}), i \in [4000, L)
\end{equation}
\begin{equation}
    P_2(x_i) = p(x_i|x_0, ..., x_{i-1}), i \in [0, 4000)
\end{equation}
After obtaining the probability of every token in the given text, we calculate the PPL:
\begin{equation}
    PPL = average(-log(P_2(x_i))), i \in [0, L)
\end{equation}

Such a calculation method can help models deliver a lower PPL compared to the former calculation method under the same test corpus. Therefore, we adopt it for LLaMA2, which only has a limited context window of 4,096, to show that a model can be good at long text language modeling even though it cannot understand long text.

\subsection{Details for Downstream Tasks}
\label{app:tasks}
We adopt QMSUM and NarrativeQA from LongBench \citep{bai2023longbench}, a comprehensive benchmark for long-text understanding, to evaluate the models' ability in long-text understanding. The average length of test cases in QMSUM and NarrativeQA are 10,614 and 18,409 respectively. Following \citet{bai2023longbench}, we use the Rouge-L score as the evaluation metric for QMSUM and the F1 score for NarrativeQA.

For the finer-grained line retrieval test, following \cite{longchat2023}, we construct a synthetic dataset, which consists of 3,200 lines of key-value pairs, and ask the model to retrieve the value of a certain key. The key-value pair is shown in the format: "A person named $\langle$key$\rangle$ has a cat. His cat's name is $\langle$value$\rangle$". The average length of each test case is 76K. We report the average retrieval accuracy of 200 runs.

The detailed prompt is: 

{\it Below is a record of lines I want you to remember. Each line begins with a person's name and its cat's name. For each line index, memorize the person's name and the cat's name. At the end of the record, I will ask you to retrieve the cat's name of a certain person. Now the record start:

A person named $\langle$key\_1$\rangle$ has a cat. His cat’s name is $\langle$value\_1$\rangle$....A person named $\langle$key\_3200$\rangle$ has a cat. His cat’s name is $\langle$value3\_200$\rangle$.

Now the record is over. I remember the cat's name of person $\langle$example\_key\_1$\rangle$ is $\langle$example\_value\_1$\rangle$.

I remember the cat's name of person $\langle$example\_key\_2$\rangle$ is $\langle$example\_value\_2$\rangle$.

I remember the cat's name of person $\langle$example\_key\_3$\rangle$ is $\langle$example\_value\_3$\rangle$.

I remember the cat's name of person $\langle$test\_key$\rangle$ is}

\end{document}